\documentclass[11pt,a4paper]{article}
\usepackage[hyperref]{acl2020}
\usepackage{times}
\usepackage{latexsym}

\usepackage{microtype}
\usepackage{amssymb}
\usepackage{amsmath}
\usepackage{listings}

\newcommand{\rev}[2]{#2}

\usepackage{subcaption}
\usepackage{url}
\usepackage{multirow}
\usepackage{array}
\usepackage[utf8]{inputenc}
\usepackage{graphicx}

\usepackage{cleveref}
\usepackage{booktabs}
\usepackage{float}
\usepackage{tipa}
\usepackage{adjustbox}

\aclfinalcopy 
\def\aclpaperid{2135} 

\title{Puzz{L}ing {M}achines: {A} {C}hallenge on {L}earning {F}rom {S}mall {D}ata }

\author{Gözde Gül Şahin\textsuperscript{$\dagger$},~
Yova Kementchedjhieva\textsuperscript{$\alpha$},~
Phillip Rust\textsuperscript{$\dagger$},~
\textbf{Iryna Gurevych\textsuperscript{$\dagger$}}\\[.3em]
	\textsuperscript{$\dagger$}Ubiquitous Knowledge Processing Lab (UKP), \\ Department of Computer Science, Technical University of Darmstadt\\
	\textsuperscript{$\alpha$}Department of Computer Science, University of Copenhagen\\
	\textsuperscript{$\dagger$}{\url{www.ukp.tu-darmstadt.de}}\\
}

\date{}

\begin{document}
\maketitle
\begin{abstract}

    Deep neural models have repeatedly proved excellent at memorizing surface patterns from large datasets for various ML and NLP benchmarks. They struggle to achieve human-like thinking, however, because they lack the skill of iterative reasoning upon knowledge. To expose this problem in a new light, we introduce a challenge on learning from small data, \textit{PuzzLing Machines}, which consists of \textit{Rosetta Stone} puzzles from Linguistic Olympiads for high school students. These puzzles are carefully designed to contain only the \textit{minimal} amount of parallel text necessary to deduce the form of unseen expressions. Solving them does not require external information (e.g., knowledge bases, visual signals) or linguistic expertise, but meta-linguistic awareness and deductive skills. Our challenge contains around 100 puzzles covering a wide range of linguistic phenomena from 81 languages. We show that both simple statistical algorithms and state-of-the-art deep neural models perform inadequately on this challenge, as expected. We hope that this benchmark, available at \url{https://ukplab.github.io/PuzzLing-Machines/}, inspires further efforts towards a new paradigm in NLP---one that is grounded in human-like reasoning and understanding.

\end{abstract}

\section{Introduction}
\label{sec:intro}
	
	\citet{kahneman2011thinking} discusses the two modes of human thinking which perfectly encapsulate the current (so called System1) and the desired state (System1+System2) of the deep learning field. System1 handles tasks that humans consider fast, intuitive and automatic, such as object detection and document classification. Recent deep learning (DL) models have shown great promise at this type of tasks---thanks to large training datasets. Yet, it is through slow, rational and sequential mechanisms that human-like abstract reasoning happens, to enable learning from just a few examples. This System2-style modeling is still in its early stages in DL, but is recognized as a much needed next step in the field~\cite{extendingML,marcus20,lecuntalk,bengiotalk}. To foster research in this promising direction, we propose a unique challenge on ``learning from small data'': \textit{PuzzLing Machines}, based on the Linguistic Olympiads---one of the 13 recognized International Science Olympiads targeted at high-school students.
	\begin{table}[!]
	\center
	\begin{adjustbox}{width=\columnwidth,center}
		\begin{tabular}{ll}
		\textbf{Chikasaw} & \textbf{English} \\
		\hline
		1. Ofi'at kowi'ã lhiyohli. & The dog chases the cat. \\
		2. Kowi'at ofi'ã lhiyohli. & The cat chases the dog. \\
		3. Ofi'at shoha. & The dog stinks. \\
		4. Ihooat hattakã hollo. & The woman loves the man. \\
		5. Lhiyohlili. & I chase her/him. \\
		6. Salhiyohli. & She/he chases me. \\
		7. Hilha. & She/he dances.  \\
		\hline
		\multicolumn{2}{c}{\textit{Now you can translate the following into Chickasaw:}} \\
		& The man loves the woman. \\
		& The cat stinks. \\
		& I love her/him. \\
		\hline
		\multicolumn{2}{c}{\textit{Translate the following into English:}} \\
		Ihooat sahollo. &  \\
		Ofi'at hilha. &  \\
		Kowi'ã lhiyohlili. &  \\
		\hline
		\end{tabular}
		\end{adjustbox}
	\caption{The ``Chickasaw'' puzzle~\cite{chikasaw}}
	\label{tab:example_puzz}
	\end{table}
    
	The \textit{PuzzLing Machines} challenge is based on one of the most common puzzle types in the Linguistic Olympiads: the \textit{Rosetta Stone} puzzles~\cite{bozhanov2013rosetta}, a.k.a. translation puzzles. An example is given in Table~\ref{tab:example_puzz}.\footnote{Copyright University of Oregon, Department of Linguistics.} Although these puzzles take the form of a traditional ``machine translation'' task, they are different in many ways: Rosetta Stone puzzles contain a minimal, carefully designed set of parallel expressions (words, phrases or sentences) in a foreign and in a familiar language (e.g., Chickasaw-English). This minimal set is \textit{just} enough to deduce the underlying translation model, which typically involves  deriving mini-grammar rules, extracting a lexicon, and discovering morphological and phonological rules. The actual task then is to translate new expressions---generally in both directions---using the model deduced from the parallel data. The assignments are carefully designed so that the expressions cannot be generated through simple analogy, but rather through the application of the discovered rules. These properties distinguish the \textit{PuzzLing Machines} challenge from the modern MT task, as it relies on deductive reasoning with linguistic concepts that are central to System2, rather than exploiting statistical properties from large datasets as in System1.

	The lack of reasoning skills of statistical systems has recently gained a lot of attention. Various datasets that require a wide range of background knowledge and different types of reasoning abilities have been introduced, such as ARC~\cite{clark2018think}, GQA~\cite{HudsonM19}, GLUE benchmarks~\cite{WangSMHLB18} and SWAG~\cite{swag}. Our challenge distinguishes from previous benchmarks with some key properties. First, most of these reasoning tasks require external scientific or visual knowledge, which makes it hard to measure the actual reasoning performance. On the other hand, our challenge does not rely on any external, multimodal or expert-level information. Second, and more importantly, \textit{PuzzLing} challenge consists of a minimal set of examples required for solution. That means, there exists no extra training data, ensuring that exploiting surface patterns would not be possible unlike in some of existing benchmarks~\cite{GururanganSLSBS18}. 
	
	In summary, this paper introduces a unique challenge, \textit{PuzzLing Machines}, made up of $\sim$100 Rosetta Stone, a.k.a translation puzzles covering 81 languages from 39 different language families based on the Linguistic Olympiads. The challenge requires System2 skills---sequential reasoning and abstraction of linguistic concepts, discussed in detail in \textsection \ref{sec:metaling}. We discuss the dataset and the linguistic phenomena in the resulting dataset supported with statistics and examples in \textsection \ref{sec:dataset}. In \textsection \ref{sec:models}, we present the results of intuitive baseline methods and strong MT baselines such as Transformers encoder-decoder \cite{VaswaniSPUJGKP17} with integrated pretrained language models as applied to these puzzles. We show that, unsurprisingly, the puzzles cannot be easily or robustly solved by currently existing methods. We hope that this benchmark is going to evoke development of new deep MT/NLP models that operate in a human-like manner and reason upon linguistic knowledge, providing a new future research direction for NLP.

\section{Meta-linguistics}
    \label{sec:metaling}
	\textit{Meta-linguistics} is defined by \citet{chomsky1976reflections} as ``the knowledge of the characteristics and structures of language'' as realised on the level of phonology, morphology, syntax and semantics. Any English speaker would likely have the linguistic capacity to produce the word \textit{undo} when asked ``What is the opposite of \textit{do}?'' Only a speaker with some level of meta-linguistic awareness, however, would further be able to reflect on the structure of the word they have produced: to identify \textit{un-} as a unit that serves to negate words, to spot its similarity in function to other units like \textit{dis-} and \textit{de-}. He/she would also be aware that \textit{un-} is not interchangeable with \textit{dis-} and \textit{de-}, since it attaches to the front of verbs and adjectives but not to nouns. 

	Meta-linguistic awareness is especially useful (and often improved) in the process of learning a new language, as it allows the learner to compare and contrast the structure and characteristics of the new language to those that he/she is already familiar with. It is desirable that systems for natural language processing possess meta-linguistic awareness, too, as that could hugely improve their cross-lingual generalizability, a problem that remains open after being approached from various engineering perspectives, often with little recourse to linguistics. However, measuring the meta-linguistic awareness of a system is not trivial. Existing probing techniques are mostly designed to measure how well neural models capture specific linguistic phenomena, e.g., whether a specific layer of an English language model can capture that \textit{undo} is negative, instead of testing for meta-linguistic awareness. Our challenge takes a step further and tests whether the model can apply the underlying morphological processes, e.g. of verbal negation through prefixing. In addition, our challenge spans a wide-range of language families and covers a variety of linguistic phenomena (see \textsection \ref{ssec:ling_phen}), that qualifies it as a favorable testbed for measuring meta-linguistic awareness. 

	Let us demonstrate how meta-linguistic reasoning skills are used to solve the ``Chickasaw puzzle'' given in Table~\ref{tab:example_puzz}. The translation model is iteratively deduced as follows: (1) the word order in Chickasaw is Subject-Object-Verb (SOV), unlike the English SVO word order; (2) nouns take different suffixes when in a subject or object position (\textit{at} and \textit{ã}, respectively); (3) verbs take a suffix for 1st person singular pronomial subject or object (\textit{li} and \textit{sa}, respectively). Notice that, crucially, it is not possible to learn the function of the prefix \textit{sa}, which corresponds to \textit{me} in English, without deducing that \textit{lhiyohli} corresponds to the verb \textit{chases} and that third person agency in Chickasaw is not explicitly expressed. As demonstrated, inferring a translation model requires iterative reasoning on the level of words, morphemes and syntactic abstractions (classes), or, to put things differently, it requires meta-linguistic awareness.

\section{The Dataset}
\label{sec:dataset}

	The puzzles from Linguistic Olympiads cover many aspects of language such as phonetics, morphology, syntax and semantics. They are carefully designed by experts according to several key criteria: (1) The puzzles should be \textit{self-contained} and \textit{unambiguous}, meaning that no prior knowledge in the foreign language is requires, just the command of one's own native language and some level of meta-linguistic awareness and that a solution is guaranteed; (2) They should require no specialized external knowledge or formal linguistic knowledge, i.e. linguistic terms are either excluded from the instructions that accompany a puzzle or they are explicitly defined; (3) The foreign language used in a puzzle should be from a truly lesser known language family (e.g. Chickasaw, Lakhota, Khmer, Ngoni), such that there is no unfair advantage to participants whose native language is related.
	
	\begin{table*}
	    \centering
	    \begin{adjustbox}{max width=\textwidth}
	    \begin{tabular}{lllll}
	        & \textbf{Language} & \textbf{Source sentence} & \textbf{Target sentence} & \textbf{Other accepted forms} \\
	        \hline
	        1.& Chickasaw & Hilha. & (She/He) dances. & She dances.  \\
	           &           &        &                  & He dances. \\ 
	        2a.& Blackfoot & Nitoki’kaahpinnaan. & We.PL2- camped. & We camped.\\
	        2b.& Blackfoot & Oki’kaao’pa. & We.PL2 camped. & We camped.\\
	        3. & Wambaya & Bardbi ga bungmanya. & The old woman ran [away]. &  The old woman ran away. \\ 
	        4. & Euskara & Umea etorri da. & The child has (come/arrived). & The child has come. \\
	            &         &                 &                               & The child has arrived.      
	    \end{tabular}
	    \end{adjustbox}
	    \caption{Examples of special transcription notation.}
	    \label{tab:notation}
	\end{table*}{}
	
	We based our data collection efforts on a rich and publicly available database of language puzzles maintained by the organizers of NACLO.\footnote{http://tangra.cs.yale.edu/naclobase/} This resource contains puzzles from IOL and a wide range of local competitions\footnote{NACLO (North America), OzCLO (Australia), UKLO (UK), Olimpíada Brasileira (Brazil), OLE (Spain), Panini (India), Russian LO, Russian Little Bear, Swedish LO, Polish LO, Estonian LO, Slovenian LO, Bulgarian LO, Netherlands LO and more.}. 
	We only included puzzles written in English (or translated to English) to ensure a quality transcription and to enable error analysis. Expert solutions are available for most puzzles; we excluded the rest. In addition to the translation puzzle type shown in Table~\ref{tab:example_puzz}, we also collected `matching' puzzles. These are two-step puzzles, in which the participants first align a shuffled set of sentences to obtain parallel data, and then translate a set of unseen sentences. We converted these puzzles to the translation puzzle format by referring to the solution files to align the training sentence pairs.
	Appendix~\ref{ssec:transc} describes how we selected the puzzles and how we transcribed them into a machine-readable format.


	The final dataset contains 96 unique puzzles from 81 languages that span 39 different language families from all over the world, as well as two creoles and two artificial languages (see Appendix~\ref{ssec:lang_list} for the full list). Some of the large language families have multiple representatives, e.g. there are 13 Indo-European languages, seven Austronesian and six from the Niger-Congo family. But the majority of languages are single representatives of their respective family. This genealogical diversity leads to a great diversity in the linguistic phenomena attested in the data. Some puzzles are designed to explore a specific aspect of the unknown language in isolation, e.g. case markers on demonstrative pronouns in Hungarian \cite{hungarian}. In general, however, the correct solution of a puzzle involves processing on the level of syntax, morphology, phonology, and semantics all at once. 

	\subsection{Linguistic Phenomena}
	\label{ssec:ling_phen}
        The foreign languages used in linguistic puzzles are purposefully chosen to demonstrate some interesting linguistic phenomena, not found in English (or in the respective source language of the puzzle) \cite{bozhanov2013rosetta}, resulting in a challenging, non-trivial translation process between these diverse languages and English. In this section, we outline some key linguistic properties of the languages found in the dataset, but the list is by no means exhaustive. 
	
		\paragraph{Syntax:} Three common configurations for the order between subject (S), verb (V) and object (O) in a sentence are exemplified in the dataset: SVO, SOV and VSO. In addition to these three, our dataset covers the rather rare OSV word order: see the example in Table~\ref{tab:dyirbal} from the Australian language Dyirbal \cite{dyirbal}.

		\paragraph{Morphology:} We see examples of highly analytic languages (e.g. Yoruba from West Africa) as well as highly polysythetic ones (e.g. Inuktitut from Canada). Within the synthetic type, we see both agglutinative languages (e.g. Turkish) and inflectional ones (e.g. Polish). Some specific morphological properties explored in the puzzles are verbal inflection with its many categories concerning tense, aspect and mood, nominal declension and noun class systems. The aforementioned ``Dyirbal'' puzzle also exemplifies an interesting classification of nouns, wherein women and dangerous animals and objects are treated as one class, men and other animals constitute another class and a third class captures all remaining nouns. The choice of the articles \textit{balan} and \textit{bagu} in Table~\ref{tab:dyirbal}, for example, is guided by this classification.

		\paragraph{Phonology:} A wide range of phonological assimilation processes interplay with the morphological processes described above and obfuscate morpheme boundaries. These can concern voicing, nasality and vowel quality, among other features. 
		As an example of morphological and phonological processes working together, consider the realization of pronomial possession in Australian language Wembawembda \cite{wembawemba}. Unlike English, which expresses this feature with pronouns \textit{his/her/its}, Wembawemba expresses it with a suffix on the noun it modifies, e.g. \textit{wutyupuk} `(his/her/its) stomach'. The form of the suffix, however, depends on the ending of the noun it attaches to and can vary greatly as shown in Table~\ref{tab:wembawemba}. 

	    \begin{table}[]
	        \centering
	        \begin{tabular}{l|llllll}
	             Form & nyuk & duk & nuk & buk & guk & uk\\
	             After & vowel & n & r & m & ng & other
	        \end{tabular}
	        \caption{Variants of a possessive suffix in Wembawemba and their phonological distribution.}
	        \label{tab:wembawemba}
	    \end{table}{}
	    
		\paragraph{Semantics:} Semantics come into play when we consider the compositionality of language and figurative speech: the phrase ``falepak hawei'' in the Indonesian language Abui, for example, literally translates into ``pistol's ear'', but a more fitting translation would be ``trigger'' \cite{abui}. 
		
		As a side note, it is important to note that while here we use extensive linguistic terminology to discuss the properties of the languages in our dataset, the high-school students who participate in Linguistic Olympiads need not and may not be familiar with any of the terminology. Their good performance depends on a well-developed meta-linguistic awareness, not on formal linguistic training.

	\subsection{Dataset statistics}
	\label{ssec:stats}

	\begin{figure}
	    \centering
	    \includegraphics[width=\linewidth]{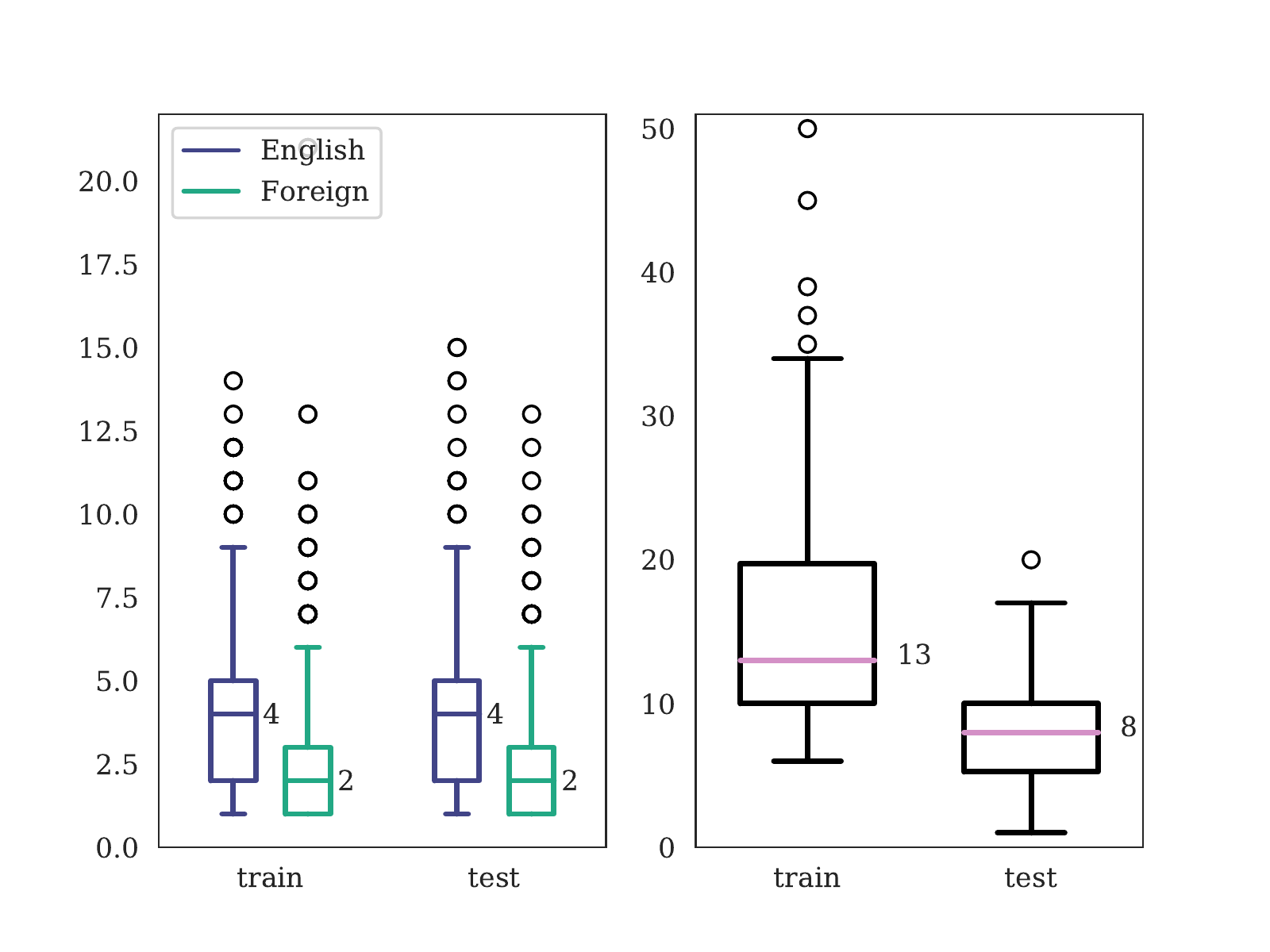}
	    \caption{Box-plots for \textbf{Left:} Word\# per language and split, \textbf{Right:} Sentence\# per split.}
	    \label{fig:sent_stats}
	\end{figure}

	In total, 2311 parallel instances are transcribed---1559 training and 752 test. 63\% of the test pairs are in the English $\rightarrow$ foreign direction, while the rest are in the foreign $\rightarrow$ English direction. 


	Statistics concerning the number of words per sentence\footnote{We naively tokenize on space.} are shown on the left of Figure~\ref{fig:sent_stats}. The majority of both training and test pairs are fairly short, but length varies considerably. This is due to the fact that some puzzles in the dataset concern the translation of individual words, some take scope over noun-modifier phrases and some, over entire sentences. English sentences are generally longer (median 4) than their translations (median 2). This is rather intuitive considering the synthetic nature of many of the foreign languages in the dataset, wherein a single long word in the foreign language may translate into 4-5 words on the English side, as in this translation from \textit{t$\Lambda$ckotoyatih} in the Mexican language Zoque  to the English \textit{only for the tooth}. 

	Sentence statistics about the length of the train and test split per problem are shown on the right of Figure~\ref{fig:sent_stats}. Intuitively, train splits are bigger than test splits. However, the number of training instances varies greatly between the puzzles, which is related to a number of factors such as the difficulty and type of the task, as well as the linguistic properties of the foreign language.

    \subsection{Train versus Test Splits}
    One property of the data splits in linguistic puzzles, which diverges from the standard paradigm in machine learning, is that the \textit{input} test data should not be considered ``held out''. On the contrary, in some cases, vocabulary items attested in the input of foreign$\rightarrow$English test instances may be crucial to the translation of English$\rightarrow$foreign test instances, and vice versa. So it is only the \textit{targets} of test instances that should be truly held out. This specificity is not ubiquitous across the puzzles, but it should be accounted for by any approach to their solution, for example by building the system vocabulary over the union of the train and input test data. 
    
\section{Baselines}
\label{sec:models}
	We attemp to solve these puzzles with models of varying complexity, i.e. from random guessing to state-of-the-art neural machine translation systems. 

	\paragraph{Random Words (RW):} Since the vocabularies of source and target languages are quite small, we test what random word picking can accomplish. We simply tokenize the training sentence pairs and then randomly choose a word from the target language's vocabulary for each token in the source sentence.\footnote{We don't use frequency of the words, i.e., pick words that occur more often, since they are not that meaningful due to the tininess of the data.}  
	
	\paragraph{FastAlign (FA):} We use the translation alignment tool FastAlign \cite{fastAlign}, to test whether the puzzles can be solved by early lexical translation models \cite{BrownPPM94}. Since FA produces alignments for each training pair, we postprocess the output to create a translation dictionary separately for each direction. We then randomly choose from the translation entries for each token in source test sentence.~\footnote{We add all aligned target phrases of the source token to the dictionary. Hence, when one target phrase is seen multiple times, it is more likely to be chosen during inference.} 

	\paragraph{Phrase Based Statistical Machine Translation (PBSMT)} Since \citet{koehn-knowles-2017-six} report that PBSMT models outperform vanilla NMT models in case of small parallel training data, we use PBSMT as one of the baselines. For the foreign$\rightarrow$English direction, we implement two models---one using no external mono-lingual English data and one otherwise.

	\subsection{Neural Machine Translation}
    We implement three different models based on Transformers \cite{VaswaniSPUJGKP17} using the implementation of \citet{ott2019fairseq}. In the first scenario, we train an off-the-shelf Transformer encoder-decoder model for each direction, referred to as \textit{Transformer}. Second, we use a strong pretrained English language model, RoBERTa \cite{roberta19}, to initialize the encoder of the NMT model for English to foreign translation. Finally, for foreign to English translation, we concatenate the translation features extracted from the last Transformer decoder layer, with the language modeling features extracted from RoBERTa \cite{roberta19}, before mapping the vectors to the output vocabulary. These models are denoted as \textit{Transformer+RoBERTa}.

\section{Experiments}
\label{sec:experiments}
	
	\subsection{Experimental Settings}
	\label{ssec:expsett}

	We first compile a subset from the puzzles that are diverse by means of languages and contain translation questions in both directions. During tuning, we use the test sentences on these puzzles to validate our models. Since our foreign languages are morphologically rich, we use BPE~\cite{bpe} to segment words into subwords. For the sentences in the foreign language, we learn the BPE from the training data, while for English sentences we use the already available GPT2-BPE dictionary to exploit English language prior. For convenience, before we train the models, we lowercase the sentences, remove certain punctuations, remove pronoun tags and brackets, and augment training data with multiple reference translations. 

	\paragraph{PBSMT:} We use Moses \cite{moses} with default settings. We employ wikitext-103 corpus to train a 5-gram English LM for the model with access to external data. The other model only uses training sentences for the LM.

	\paragraph{NMT:} Following the suggestions for low-resource NMT systems by \citet{SennrichZ19}, we use small and few layers and high dropout rates. Similarly we use the smallest available language model (RoBERTa Base) and freeze its parameters during training to reduce the number of trainable parameters. We tune the following hyper-parameters: BPE merge parameter, learning rate and number of epochs. 

	\begin{figure*}[!]
	      \centering
	       \begin{subfigure}[b]{0.47\textwidth}
	          \includegraphics[width=\textwidth]{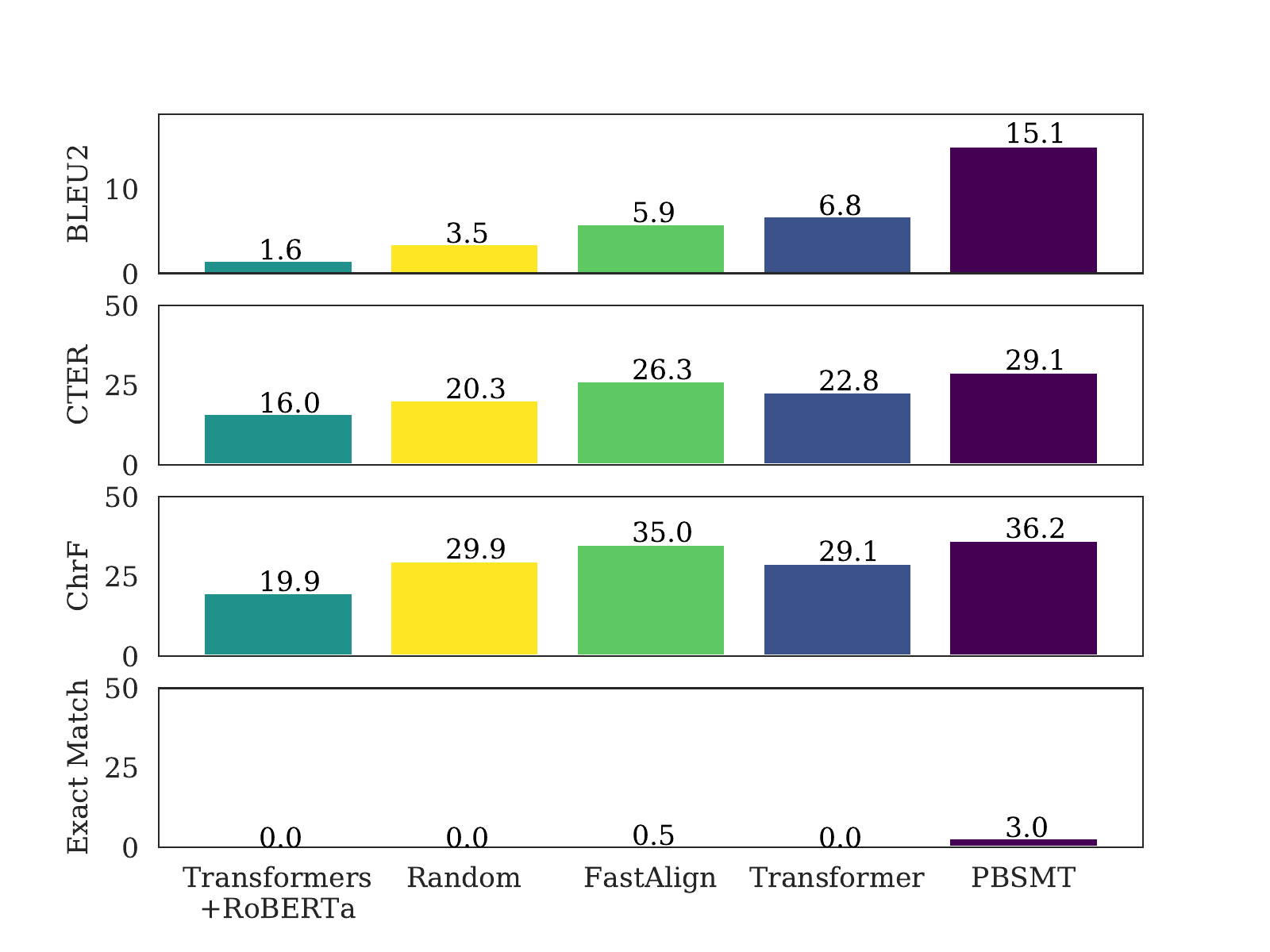}  
	      \end{subfigure}
	      \begin{subfigure}[b]{0.47\textwidth}
	          \includegraphics[width=\textwidth]{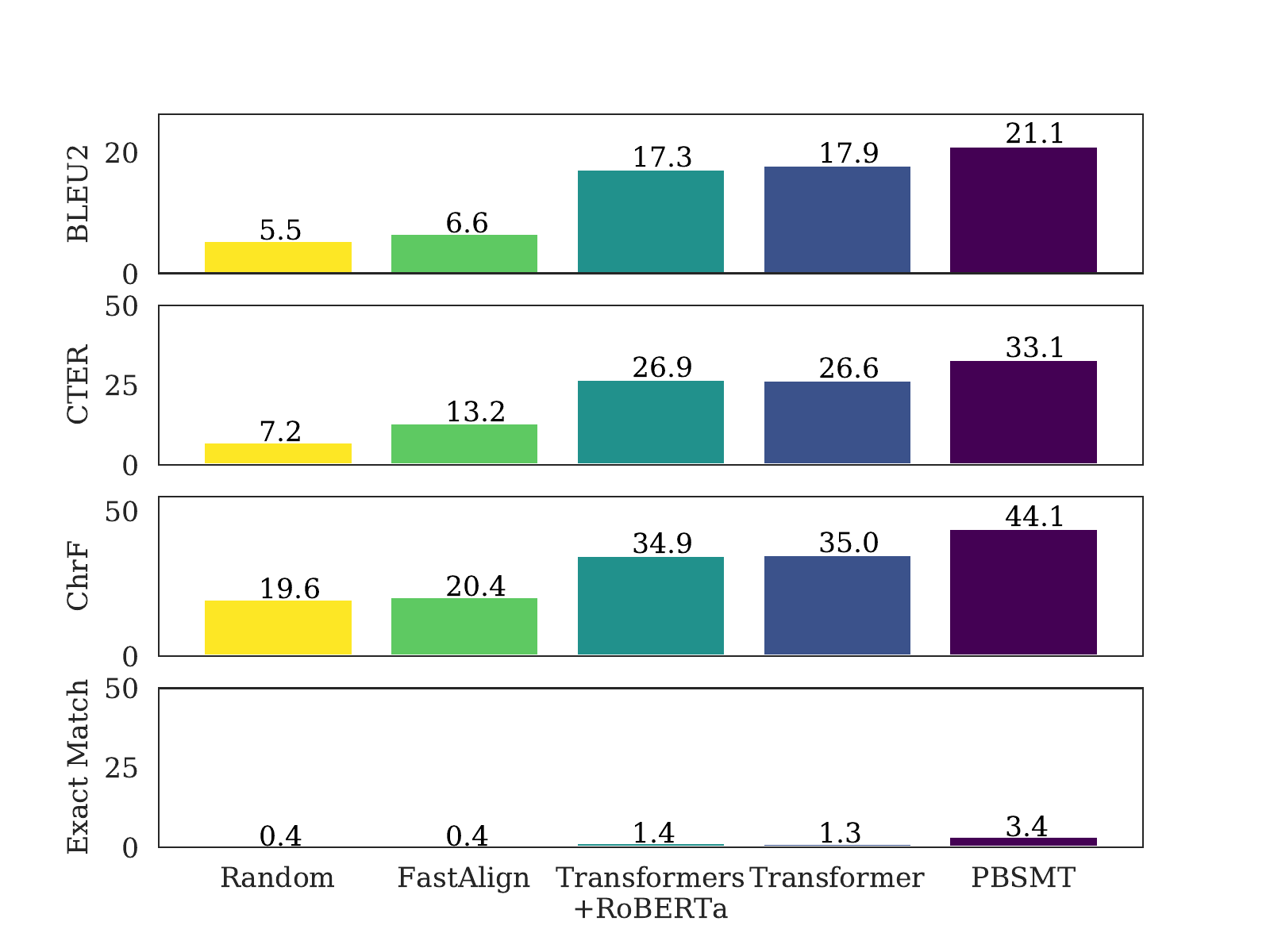} 
	      \end{subfigure}
	\caption{Main results (best viewed with color). \textbf{Left:} English$\rightarrow$foreign \textbf{Right:} foreign$\rightarrow$English.}
	\label{fig:main_res}
	\end{figure*}
	
	\subsection{Evaluation Metrics}
	\label{sec:eval}
		The submissions to Linguistic Olympiads are manually graded by experts. For a full mark, an exact solution has to be provided, as well as a correct and detailed discussion of the underlying processes that led to this solution, e.g., concerning findings about word-order, the function of individual morphemes, etc. Participants are also given partial marks in case of partial solutions or valid discussions. Since we don't have access to expert evaluation, we use readily available automatic machine translation measures. We also note grading of system interpretations or its solution steps as an interesting future research direction.

		The first is the BLEU \cite{bleu} score since it is still the standard metric in MT. We use BLEU-2 to match the lower median of sentence lengths we observe across the English and the foreign data (see Fig~\ref{fig:sent_stats}). BLEU matches whole words rather than word pieces, which prevents us from assigning partial credit to subword matches, which could be especially relevant for foreign target languages with rich morphology. We therefore use three additional metrics that operate on the level of word pieces: CharacTER \cite{characTER}, ChrF \cite{popovic16} and ChrF++ \cite{popovic17}. CharacTER is a measure derived from TER (Translation Edit Rate), where edit rate is calculated on character level, whereas shift rate is measured on the word level. It calculates the minimum number of character edits required to adjust a hypothesis, until the reference is matched, normalized by the length of the hypothesis sentence. For easier comparison, we report $1.0-characTER$ scores. ChrF is a simple F-measure reflecting precision and recall of the matching character n-grams. ChrF++ adds word unigrams and bi-grams to the standard ChrF for a higher human correlation score. We experiment with different combinations of character n-grams ($n=3,5$ as suggested in \citet{popovic16}) and word n-grams ($n=0,1,2$ as suggested in \citet{popovic17}).
		
		Finally, we also measure the average exact match of the puzzles, which is calculated as 1 if the prediction and reference sentences match and 0 otherwise. 
	    As it is not feasible to report and compare results on all of these metrics (nine in total), we compute the pair-wise Pearson correlation coefficient between them, and average over all pairs to arrive at the following four metrics that show the least correlation with each other: BLEU$-2$, CharacTER, ChrF$-3$ and exact match. We note, however, that of these four, exact match is really the most meaningful metric. Since the sentences in the dataset are rather short and the puzzles are designed to be solvable and unambiguous, an exact match should be attainable. Moreover, as the puzzles in the dataset are of varying difficulty, the average exact match score can be seen as a continuous metric. 
	
\section{Results and Analysis}
\label{sec:res}

	We report the results for the best models in Fig.~\ref{fig:main_res}. The hyperparameter configuration and the development set results are given in Appendix~\ref{ssec:hyperparams}. The maximum exact match score among all results is only $3.4$\%; and the highest scores are consistently achieved by PBSMT models on both directions and dataset splits. 

	The overall results for foreign $\rightarrow$ English are generally higher than English $\rightarrow$ foreign. This may be due to (a) having longer sentences for English; (b) the scores (except from EM) being more suitable for English (even the character-based ones) or (c) the more challenging nature of translation into foreign languages, which needs another dedicated study.   

	\paragraph{English$\rightarrow$Foreign:} Initializing the NMT encoder with RoBERTa has severely worsened the results, compared to standard Transformer model. We believe the main reason is the imbalance between encoder (huge encoder) and the decoder (tiny decoder), that makes training very challenging. The gap between the simplest baselines (RW, FA) and more sophisticated models (Transformers, PBSMT) is also considerably small; FA even surpassing Transformers's CTER and ChrF performance. For most of the foreign languages, even when two words are semantically distant, there may still be significant morpheme overlap. These suggest that simple lexical alignment models (including random assignment) can achieve higher \textit{partial} matching scores that hints at the unreliability of CTER and ChrF measures for the puzzles. 

	\paragraph{Foreign$\rightarrow$English:} We observe that the gap between the simple and more sophisticated baselines are higher in this direction by means of all measures, as we would expect. Using RoBERTa features in the decoder does not hurt the performance while providing a small increase in EM score compared to standard Transformers. It should be noted that the decoder is still tiny and LM features are only incorporated via a separate linear layer at a very late stage, which prevents the imbalance problem we saw in English $\rightarrow$ foreign. 

	We see similar results for the validation data with the exception that Transformer-based models achieve either higher or the same EM scores than PBSMT while surpassing PBSMT's BLEU-2 scores in foreign $\rightarrow$ English. It supports the findings of \citet{SennrichZ19}, drawing attention to the importance of hyper-parameter tuning for low-resource NMT models. 

		\begin{table*}
		\centering
		\resizebox{\linewidth}{!}{%
			\begin{tabular}{lll|ll}
			& \textbf{Chikasaw} & \textbf{English} & \textbf{PBSMT} & \textbf{Transformer}\\
			\hline
			\multicolumn{3}{c}{\textit{Now you can translate the following into Chickasaw:}} & & \\
			(1) & \textcolor{olive}{Hattakat ihooã hollo.} & The man loves the woman. & the the woman hattakã hollo & ihooat hattakã hollo \\
			(2) & \textcolor{olive}{Kowi'at shoha.} & The cat stinks. & the lhiyohli stinks & ofi'at shoha\\
			(3) &\textcolor{olive}{Holloli.} & I love her/him. & i love him & lhiyohlili \\
			\hline
			\multicolumn{3}{c}{\textit{Translate the following into English:}} \\
			(4) & Ihooat sahollo. & \textcolor{olive}{The woman loves me}. & ihoothe sahollo & the woman loves the man\\
			(5) & Ofi'at hilha. & \textcolor{olive}{The dog dances.} & the(he/she) dances & the cat chases the dog \\
			(6) & Kowi'ã lhiyohlili. & \textcolor{olive}{I chase the cat.} & cat ch thei chase (him/her) & the dog stinks \\
			\hline
			\end{tabular}%
		}
		\caption{Predictions for the ``Chickasaw'' puzzle. Gold-standard target sentences are shown in yellow.}
		\label{tab:example_puzz_pred}
		\end{table*}
		
	\subsection{Error Analysis}
	\label{sec:err_analysis}
		We perform manual error analysis on the predictions of our top two models for the Chickasaw puzzle presented in Table~\ref{tab:example_puzz}. The predicted translations are shown in Table~\ref{tab:example_puzz_pred}. We also provide the predictions of the simple baselines in Appendix~\ref{ssec:chick_add} for comparison. Although the PBSMT model is best on average, we find that for this particular puzzle, the Transformer model did much better. PBSMT had very few hits overall: it correctly chose to include the lexical items \textit{hattak} and \textit{hollo} in (1), but the position and inflection of the former is incorrect. In (5) and (6) there are indications of correct lexicon induction, but the overall quality of the translations is very poor both in terms of accuracy and fluency. The Transformer model, on the other hand, predicts fluent translations in both directions. In the direction from English to Chickasaw, we see that the model correctly acquired the relevant morphological patterns: subjects take suffix \textit{at}, objects take suffix \textit{ã}, and, importantly, that first person agency is expressed through suffix \textit{li}. The translations are still not perfect, though, due to lexical confusion: the words for \textit{cat} and \textit{dog} have been swapped in both (1) and (2), as well as the words for \textit{love} and \textit{chase} in (3). In the direction from Chickasaw to English, the Transformer's predictions remain fluent, but they hardly relate to the input. Contrary to the overall results, for this puzzle translation \textit{to} English appears to be more challenging for the model.

\section{Related Work}
\label{sec:relwork}

	Recently, reasoning tasks and datasets that require natural language processing have been introduced, such as common-sense reasoning in the form of pronoun resolution e.g., WSC~\cite{wsc}, multiple-choice question answering e.g., SWAG~\cite{swag} and ARC~\cite{clark2018think}; inference tasks in the form of binary or multi-label classification problems e.g., the GLUE benchmarks~\cite{WangSMHLB18}; and visual reasoning in the form of question answering~\cite{ZellersBFC19} e.g., GQA~\cite{HudsonM19}. In these tasks, the required level of semantics is mostly limited to single sentences rather than a collection; almost all tasks target  English; data is derived from running text and is mostly close-domain. In addition, some require external knowledge bases or high-level knowledge on physical models or experiments as in ARC classified by \citet{BoratkoPMYDMCFK18}, which leaves room for accumulating errors from external parts and complicates the analysis of individual parts like reasoning. 

	Another body of early work on symbolic AI provides a different set of tools to model reasoning such as rule-engines, rule-induction algorithms, logic programs and case-based reasoning models~\cite{Kolodner92}. However, it is not trivial to represent and model our task in these frameworks, since they mostly require defining primitives, expressions, discrete features and cases. Furthermore, the strength of statistical/neural models has been repeatedly shown to surpass rule-based models. Our goal is to encourage researchers to incorporate reasoning into statistical models, rather than replacing them with symbolic models.

\section{Conclusion and Future Work}
\label{sec:conc}

  	The field of NLP has developed deep neural models that can exploit large amounts of data to achieve high scores on downstream tasks. Still, the field lacks models that can perform human-like reasoning and generalization. To mitigate this gap, we draw inspiration from the \textit{Linguistic Olympiads} that challenge the meta-linguistic and reasoning abilities of high-school students. We create a new benchmark dataset from available Linguistic Puzzles that spans over 81 languages from 39 language families, which is released at \url{https://ukplab.github.io/PuzzLing-Machines/}. We implement and evaluate simple baselines such as alignment, and state-of-the-art machine translation models with integrated a pretrained English language model. We show that none of the models can perform well on the puzzles, suggesting that we are still far from having systems with meta-linguistic awareness and reasoning capabilities. 

\section{Acknowledgements}
\label{sec:ackl}
This work was supported by the German Research Foundation through the German-Israeli Project Cooperation (DIP, grant DA 1600/1-1 and grant GU 798/17-1). We would like to thank Liane Vogel, Marc Simon Uecker and Siddharth Singh Parihar for their great help during the project. We are grateful to Dragomir Radev for his feedback and continuous help with encoding problems encountered during puzzle transcription. We thank Adam Lopez and Ilia Kuznetsov for providing feedback on early drafts of the paper. We thank the area chairs and the senior area chair, whose comments helped us improve the paper. 

\bibliography{acl2020}
\bibliographystyle{acl_natbib}

\clearpage
\appendix

\section{Appendices}
\label{sec:appendix}

	\subsection{\rev{}{Transcription of Puzzles}}
	\label{ssec:transc}
	
	The puzzles are generally provided as pdf files. Many languages in the dataset use the Latin script, optionally with some diacritics. Some which use a non-Latin script (or have no writing system at all), are transcribed with IPA or transliterated into the Latin script. Only one language, Georgian, uses a non-Latin script, namely the Mkhedruli script.
	As there are various types of puzzles presented at the Olympiads, we identified the ones relevant to our task through automatic filtering for the keywords ``translation'' or ``matching'', and manually verified the results. 
	
	To represent linguistic puzzles in a unified, machine-readable format, we defined a JSON format shown in Appendix~\ref{ssec:json_file}. The relevant data was manually extracted from the PDF files and mapped to this format in a semi-automated fashion. We faced encoding issues with many of the puzzles. For some of these, the database owner kindly provided us with the source files of the pdf documents, which enabled us to generate UTF-8 encoding of the data; others we fixed manually. Some puzzles, which use pictorial
	scripts or are otherwise UTF-8 incompatible, were discarded.

	During the transcription we came across various formats of linguistic annotation in the puzzles. This kind of information was not consistently provided across puzzles, but we included it where available, as it can be both helpful and crucial to a correct solution. 
	In the next paragraphs, we provide details on the different types of annotated information and the standardized format we used to encode those. 
	
	\paragraph{Gender distinction in pronouns:} When the foreign language does not mark gender on pronouns (or omits pronouns altogether), singular pronouns in the English translations are provided consistently as \textit{(he/she)} and \textit{(him/her)}, or \textit{(he/she/it)} and \textit{(his/her/its)}, as in Ex.~1 in Table~\ref{tab:notation}. During evaluation, instances of this notation are accepted, as well as instances of the individual alternatives.

	\paragraph{Number marking on pronouns:} When the foreign language marks two levels of plurality for the second person pronoun \textit{you}, 
	they are marked accordingly as \textit{you.SG} and \textit{you.PL}. Some languages make a distinction between plural forms concerning two objects and plural forms concerning three or more objects. In this case, we mark pronouns (not just \textit{you}, but also \textit{we} and \textit{they}) with the notation \textit{.PL2} and \textit{.PL3}, respectively. Some languages also make a distinction between an inclusive \textit{we} `you and me' and an exclusive \textit{we} `me and someone else'. We reserve \textit{we.PL2} for the inclusive sense, and mark the exclusive sense with \textit{we.PL2-}. See examples 2a and 2b in Table~\ref{tab:notation}.
	The notation presented here holds for both personal pronouns, e.g. \textit{you}, and possessive pronouns, e.g. \textit{your}. During evaluation, we disregard this notation on the side of the target language.  

	\paragraph{Zero-to-one matching:} Words that are semantically implied or required by English grammar, but not directly expressed on the side of the foreign language are shown in square brackets in some of the puzzles, as in Table~\ref{tab:notation}-Ex.~3. This bracketing exists only to aid the learning of a translation model. During evaluation, we remove these brackets from the target test sentences.

	\begin{table*}[ht!]
        \centering
        \begin{adjustbox}{max width=\linewidth}
        \begin{tabular}{lllllll}
            \textbf{Source} & balan & waymin & bambun & baNgu & jugaNgu & jamiman.\\
            \textbf{Gloss} & OBJ & mother-in-law & healthy&  SUBJ & sugar-SUBJ & fat-MAKE.\\
            \textbf{Target} & \multicolumn{6}{l}{Sugar makes the healthy mother-in-law fat.}\\
        \end{tabular}
        \end{adjustbox}
        \caption{Example sentence in Dyibral.}
        \label{tab:dyirbal}
    \end{table*}{}
    
	Notice that number marking and special notation for zero-to-one matching is not ubiquitous across the puzzles. We included it only when it was provided in the original puzzle.

	\paragraph{Multiple reference translations:} Occasionally, several possible translations are listed in a puzzle for a given word, phrase or sentence--see Table~\ref{tab:notation}-Ex.~4. We represent these options inside parenthesis separated with a slash (/), e.g., (alternative1/.../alternative N). Since the alternatives are of different granularity, nested bracketing may sometimes occur. During evaluation, we calculate the scores between the prediction and all possible references, and take the maximum.

	\paragraph{Additional information} 
	Roughly half of the puzzles contain remarks on individual characters and diacritics in the inventory of the foreign language, e.g. ``In the following orthography a colon (:) marks a long vowel, and the \textipa{P} symbol marks a glottal stop (like the sound in the middle of uh-oh)''. In many cases, the instructions state that these are pronunciation notes, i.e. they are made available only to allow the participants to vocalize the words they see on the page. On some occasions, however, they might introduce a character that is not present in the training data, but is relevant to the translation of the test sentences, e.g. the voiceless counterpart of a voiced consonant in a language with voice assimilation. As this kind of information cannot be mapped to the parallel data format, we include it in a separate field in the JSON files, directly as it appeared in the puzzles. \footnote{We believe that even if all instances of such remarks are ignored, the puzzles should remain mostly solvable, but we note that without this information, the ceiling for human performance would not be quite 100 percent.} 
	
	With the aforementioned guidelines, each puzzle was transcribed by one transcriber and verified by at least one other transcriber. 
	For the test pairs, the direction of translation is stored as well, since a possible and singular solution is only guaranteed in the direction as given in the puzzle. 



\subsection{JSON File Format}
\label{ssec:json_file}
    Each puzzle is represented with a JSON file containing the following fields: \textsc{source\_lang}, \textsc{target\_lang}, \textsc{meta}, \textsc{train} and \textsc{test}. Each field is explained in Table~\ref{table:json_fields}. 

	   \begin{table*}[!ht]  
	      \centering
	      \small
	        \begin{tabular}{|p{2cm}|p{4cm}|p{6cm}|} 
	            \hline
	            \textbf{Field} & \textbf{Definition} & \textbf{Example} \\
	            \hline
	            \textsc{source\_lang} & Name of the source language & Foreign language e.g., Kiswahili, Pali \\
	            \hline
	            \textsc{target\_lang} & Name of the target language & English \\
	            \hline
	            \textsc{meta} & Additional information about the foreign language provided in the original puzzle (as free text) & "The sound represented as ã is a 'nasalized' vowel. It is pronounced like the 'a' in 'father', but with the air passing through the nose, as in the French word 'ban'." \\
	            \hline
	            \textsc{train} & Parallel training sentences given as a list of lists & [[``Bonjour'', ``Good morning''], [``chat'', ``cat'']], where the source and the target language is French and English respectively. \\
	            \hline
	            \textsc{test} & Parallel test sentences with direction information & [[``Bonjour'', ``Good morning'', \textgreater], [``chat'', ``cat'', \textless]]. ``\textgreater'' implies that the translation is required from source to target language, vice versa for ``\textless'' \\
	            \hline
	        \end{tabular}
	        \caption{JSON file format used in the linguistic puzzles shared task}
	        \label{table:json_fields}
	    \end{table*}

\subsection{Development Results}
\label{ssec:add_res}

\begin{figure*}[!ht]
          \centering
           \begin{subfigure}[b]{0.47\textwidth}
              \includegraphics[width=\textwidth]{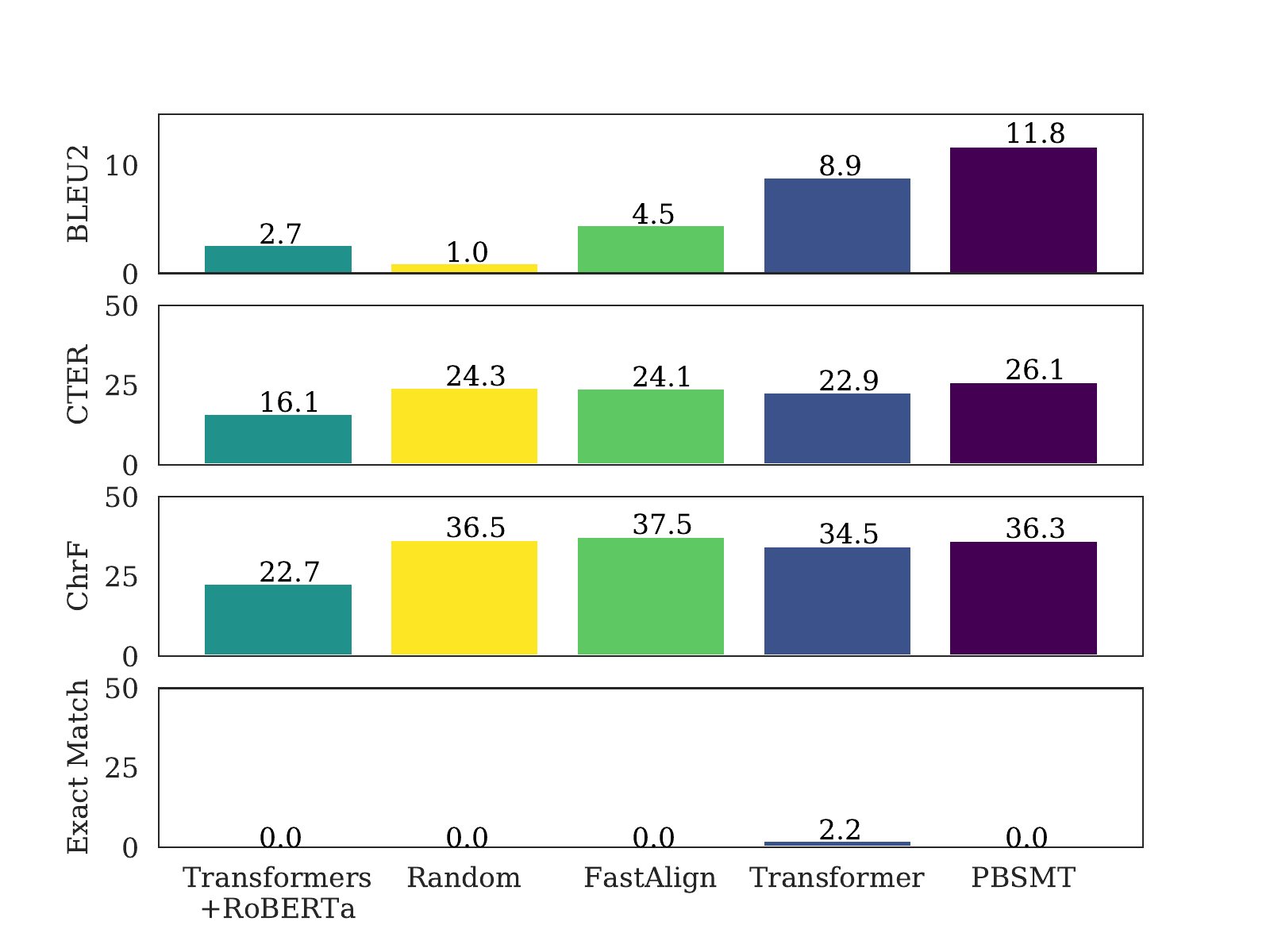}  
          \end{subfigure}
          \begin{subfigure}[b]{0.47\textwidth}
              \includegraphics[width=\textwidth]{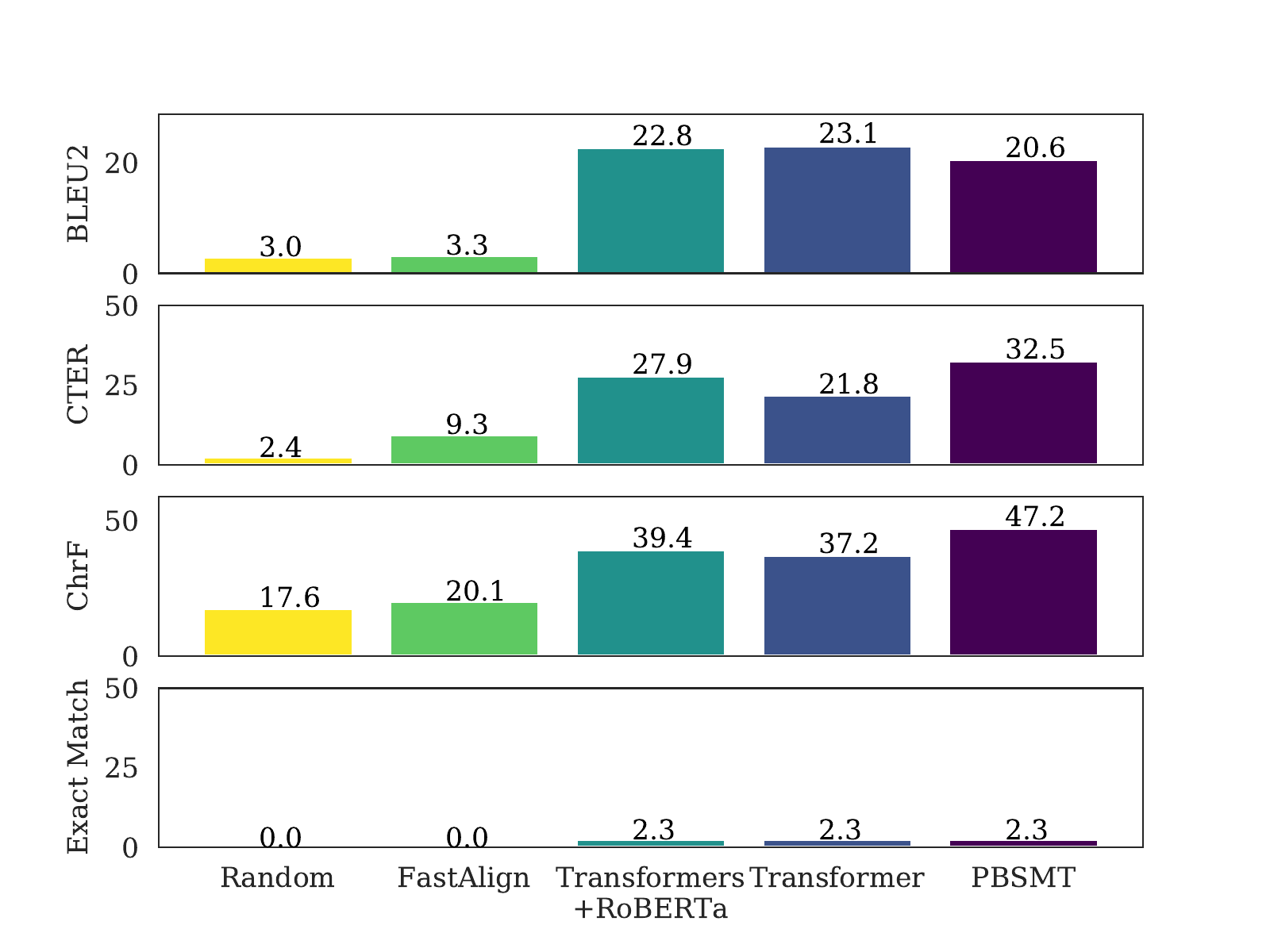} 
          \end{subfigure}
    \caption{Development set results. \textbf{Left:} English$\rightarrow$foreign \textbf{Right:} foreign$\rightarrow$English}
    \label{fig:dev_res}
    \end{figure*}

    The results on the validation set are given in Fig.~\ref{fig:dev_res}.

\subsection{Hyperparameter Settings}
\label{ssec:hyperparams}
    The best hyperparameters found for each NMT model is given as following. \textit{FA:} word to word alignments; PBSMT for English$\rightarrow$Foreign: word alignment with external English LM; PBSMT for Foreign$\rightarrow$English: BPE with 30 merge operations. For both Transformers-based models in Foreign$\rightarrow$English direction, we used BPE with 10 merge operations, learning rate of 0.001 and 500 epochs; while for the standard Transformer in English$\rightarrow$Foreign direction, BPE with 30 merge operations have been used. For all models except from Transformers with RoBERTa encoder, both the encoder and decoder had 1 layers, and all hidden dimesions were set to 128, dropout was set to 0.3, and the models were trained with Adam optimizer. For Transformer with RoBERTA LM Encoder for English$\rightarrow$Foreign, we have used 0.0001 learning rate with reduction on plateau, batches of size 2, dropout of 0.1, 1 layer, 64 embedding units, 128 hidden units, and BPE with 5 merge operations.

\subsection{Chickasaw Additional Predictions}
\label{ssec:chick_add}

    In Table~\ref{tab:baseline_pred}, the predictions of RW and FA are shown for comparison. 
    \begin{table*}
    \centering
    \resizebox{\linewidth}{!}{%
        \begin{tabular}{lll|ll}
        & \textbf{Chikasaw} & \textbf{English} & \textbf{RW} & \textbf{FA}\\
        \hline
        (1) & \textcolor{olive}{Hattakat ihooã hollo.} & The man loves the woman. & Ihooat lhiyohli hollo salhiyohli ofi'at. & The hollo loves the woman. \\
        (2) & \textcolor{olive}{Kowi'at shoha.} & The cat stinks. & Lhiyohlili lhiyohlili kowi'ã. & The lhiyohli shoha. \\
        (3) &\textcolor{olive}{Holloli.} & I love her/him. & Ofi'ã hilha lhiyohlili. & I love lhiyohlili. \\
        \hline
        (4) & Ihooat sahollo. & \textcolor{olive}{The woman loves me}. & Dog loves & Ihooat sahollo \\
        (5) & Ofi'at hilha. & \textcolor{olive}{The dog dances.} & I the &  ofi'at he dances \\
        (6) & Kowi'ã lhiyohlili. & \textcolor{olive}{I chase the cat.} & stinks cat & Kowi'ã I chase (him/her). \\
        \hline
        \end{tabular}%
    }
    \caption{Predictions of the simple baseline models for the ``Chickasaw'' puzzle. Gold-standard target sentences are shown in yellow.}
    \label{tab:baseline_pred}
    \end{table*}

\subsection{List of Languages and Families}
\label{ssec:lang_list}
    The full list for the languages and the families they belong to, as classified in WALS \cite{wals} and, where WALS lacks an entry, Glottolog \cite{glottolog}, are given in Table \ref{tab:lang_list}.

    \begin{table*}[!ht]
    \center
        \begin{tabular}{|l|l||l|l|}
            \hline
            \textbf{Language} &  \textbf{Family} & \textbf{Language} & \textbf{Family} \\ \hline 
            Abkhaz & Northwest Caucasian & Luiseño & Uto-Aztecan \\ \hline 
            Abma & Austronesian & Madak & Austronesian \\ \hline 
            Abui & Timor-Alor-Pantar & Malay & Austronesian \\ \hline 
            Afrihili & Artificial & Maori & Austronesian \\ \hline 
            Amele & Trans-New Guinea & Mayangna  & Misumalpan \\ \hline 
            Ancient Greek & Indo-European & Miwoc & Penutian \\ \hline 
            Bambara & Mande & Muna & Austronesian \\ \hline 
            Basque & Basque & Nahuatl & Uto-Aztecan \\ \hline 
            Beja & Afro-Asiatic & Ndebele & Niger-Congo \\ \hline 
            Benabena & Trans-New Guinea & Nen & Trans-New Guinea \\ \hline 
            Blackfoot & Algic & Nepali & Indo-European \\ \hline 
            Bulgarian & Indo-European & Nhanda & Pama-Nyungan \\ \hline 
            Central Cagayan Agta & Austronesian & Norwegian & Indo-European \\ \hline 
            Chamalal  & Nakh-Daghestanian & Nung & Tai-Kadai \\ \hline 
            Chickasaw & Muskogean & Old English & Indo-European \\ \hline 
            Choctaw & Muskogean & Pali & Indo-European \\ \hline 
            Cupeño & Uto-Aztecan & Papiamento & creole \\ \hline 
            Danish & Indo-European & Persian & Indo-European \\ \hline 
            Dyirbal & Pama-Nyungan & Polish & Indo-European \\ \hline 
            Esperanto & Artificial & Proto-Algoquian & Algic \\ \hline 
            Fula & Niger-Congo & Quechua & Quechuan \\ \hline 
            Georgian & Kartvelian & Somali & Afro-Asiatic \\ \hline 
            Guaraní & Tupian & Swahili & Niger-Congo \\ \hline 
            Haitian Creole & Creole & Tadaksahak & Songhay \\ \hline 
            Hmong & Hmong-Mien & Tanna & Austronesian \\ \hline 
            Hungarian  & Uralic & Teop & Austronesian \\ \hline 
            Icelandic & Indo-European & Tok Pisin & creole \\ \hline 
            Ilokano & Austronesian & Tshiluba & Niger-Congo \\ \hline 
            Inuktitut & Eskimo-Aleut & Turkish & Altaic \\ \hline 
            Irish & Indo-European & Udihe & Altaic \\ \hline 
            Jaqaru & Aymaran & Waanyi & Garrwan \\ \hline 
            Kabardian & Northwest Caucasian & Wambaya & Mirndi \\ \hline 
            Kayapo & Macro-Ge & Warlpiri & Pama-Nyungan \\ \hline 
            Kimbundu & Niger-Congo & Welsh & Indo-European \\ \hline 
            Kunuz Nubian & Eastern Sudanic & Wembawemba & Pama-Nyungan \\ \hline 
            Kurdish & Indo-European & Witsuwit’en & Dené–Yeniseian \\ \hline 
            Lakhota & Siouan & Yidiny & Pama-Nyungan \\ \hline 
            Lalana Chinantec & Oto-Manguean & Yolmo & Sino-Tibetan \\ \hline 
            Latvian & Indo-European & Yonggom & Nuclear Trans New Guinea \\ \hline 
            Lopit & Nilo-Saharan & Yoruba & Niger-Congo \\ \hline 
             &  & Zoque & Mixe-Zoque \\ \hline 
    \end{tabular}
    \caption{Full list of languages and their families.}
    \label{tab:lang_list}
    \end{table*}

\end{document}